\documentclass[runningheads]{llncs}

\usepackage[T1]{fontenc}
\def\doi#1{\href{https://doi.org/\detokenize{#1}}{\url{https://doi.org/\detokenize{#1}}}}
\usepackage{graphicx}
\usepackage{subfig}

\usepackage{listings}
\begin{document}
\title{Anomaly Detection for Fraud in Cryptocurrency Time Series}
%
%
\author{Eran Kaufman\inst{1} \and
Andrey Iaremenko \inst{3}}
\authorrunning{E. Kaufman}
%
\institute{Ben Gurion University of the Negev, Israel\and
HUB Security, Israel}
\maketitle              
\begin{abstract}
Since the inception of Bitcoin in 2009, the market of cryptocurrencies has grown beyond
 initial expectations as daily trades exceed \$10 billion.
As industries become automated, the need for an automated fraud detector becomes very apparent.
 Detecting anomalies in real time prevents
  potential accidents and economic losses. Anomaly detection in multivariate time 
  series data poses a particular challenge because it requires simultaneous 
  consideration of temporal dependencies and relationships between variables.
  Identifying an anomaly in real time is not an easy task specifically because of the 
  exact anomalistic behavior they observe.
  Some points may present pointwise global or local
  anomalistic behavior, while others may be anomalistic due to their frequency or seasonal behavior
   or due to a change in the trend.
  In this paper we suggested working on real time series of trades of Ethereum from specific accounts
  and surveyed a large variety of different algorithms traditional and new.
  We categorized them according to the strategy and the anomalistic behavior which they search 
  and showed that when bundling them together to different groups, they can prove to be 
  a good real-time detector with an alarm time of no longer than a few seconds and with very high confidence.
\end{abstract}
\section{Introduction}\label{sec1}
Blockchain technology has become pervasive today in many industries, such as the financial industry,
 cloud infrastructure, supply chain management and others. 
 Blockchain is a distributed decentralized ledger with guarantee of immutability 
 of the data stored in the ledger across multiple computational nodes.
  This cryptography-based technology removes the central authority and enables fast
   and trusted sharing of data and exchange of value  directly between two or more parties.
    Today there are many kinds of blockchains, customized to specific needs and environments.
     In the monetary sector there are multiple cryptocurrencies traded on the open markets,
 just like a regular forex.
  Moreover, in the US, Japan and Europe cryptocurrency trading is officially regulated.
Blockchain based payments with cryptocurrencies (e.g. Bitcoin, Ethereum) provide fast,
 cheap, anonymous and secure payments worldwide. 
 However, the immutability of the payment transaction presents a problem when a malicious actor obtains
  control of the access to the Bitcoin account and moves the tokens to another account on the blockchain.
   In particular, it is practically impossible to rewind the transaction and retrieve the money.
    Thus the largest security issue with cryptocurrencies is the access to the private key
     (a random number of 32 bytes), which represents the entire monetary holdings of a personal account on the blockchain.
 This risk is exponentially higher in the case of institutions handling large amounts of money and managing third party
  assets in an automated and round-the-clock manner.
   Therefore automated malicious activity detection mechanisms are required to keep pace
    with the organizational use of blockchain. 
For the financial use of blockchain there are relatively few parameters and data fields
 to inspect with automated anomaly detection algorithms, similar to regular banking fraud detection.
For Bitcoin transactions, the following fields are most relevant for anomaly detection:
\begin{enumerate}
\item	Payment amount. Each account has relatively unique behavior patterns
 regarding transaction amounts as stand-alone metric (like groups of values)
  and also relative to the time scale (time of day, weekly and monthly scales).
\item	The destination address. Amount value behavior is coupled to the recepient address field.
 There may be a small number of addresses that receive similar payments over time, 
 and one-time addresses with a single payment.
\item	The Fee value. Fee is a transactional surcharge for the miners
 (or maintainers) of the blockchain infrastructure. Bitcoin fees are traditionally
  predefined in the wallet application. 
\end{enumerate}
For Ethereum transactions, the following fields are most relevant to anomaly detection:
\begin{enumerate}
\item	The payment amount value, similar to Bitcoin.
\item	The destination address, similar to Bitcoin. 
\item	Gas limit and Gas price. These two parameters reflect the fee paid 
to the  maintainers of the blockchain infrastructure.
\end{enumerate}
 In ethereum the fee is calculated in real time for each transaction, 
 based on the current network load and on the required speed 
 of transaction approval on the blockchain. 
In the Ethereum network, a transactional request is an actual computation in a dedicated
 virtual machine in the network. Gas is the unit of work to execute 
 specific computational operations on the blockchain. Each transaction has a base fee to
  be included in the blockchain, based on the complexity of the required calculations. 
  The transaction initiator can pay an additional fee to secure faster processing 
  and approval time. Thus the fee component could be an indicator of account fraud.
All of the above parameters are trackable per account and per unit of time 
for use in anomaly detection processes. Each account has a different behavior 
over time and in between temporal epochs. 
So the anomaly detection methods should combine training and inference models.
 The algorithms seek anomalies in the spending behavior of the account over
  a period of time, for example a comparably large transaction amount or a group of
   many small transactions. 
   
   \subsubsection{Our contribution.}
   We focus our attention on the spending behavior of an account over time. Unlike previous work
   in this field which focused on documented Ponzi schemes, our aim is to find discrepancies in 
   an online working account.
  We conducted comparative research using multiple anomaly detection algorithms from
    a vast range of algorithms and models
    and compared their results in order to find the best suited algorithms for both researchers and practitioners alike.

    The algorithms must be both very reliable, and also run in real time, meaning the detection alarm must be raised within  a few seconds of 
    its occurrence.
    Since we work on unsupervised data, 
    we found that `hand labeling' anomalies rarely works well.
    This is because the question often posed to an arbiter -- when is an anomaly really an anomaly? -- is not
     a simple one to answer.
    For this reason we ran a large scale of algorithms from different approaches
     and categorized them into different categories.
    An anomaly 
      was decided to be an anomaly only when the majority out of 
     multiple algorithms from the
    \emph{same} category made the same decision.
    (We found that deciding anomalies based on different categories led to poor results.)
%
    Secondly, In regard to the feature space, out of the many possible features,
    we identified three features as being the most significant: These are payment amount, Gas price, and gas limit
     (as explained above).
     Moreover, we did both a univariate and a multivariate analysis of the data,
     i.e. we referred to the vector of features both as individual time series as well a combined three-vector.
    By doing so we mark the fact that while each feature may not be ''out of range'' individually,
     the vector as a whole can be anomalous as, for example, the gas price and gas limit should be correlated.
     
     We combined both pointwise anomalies based on representations of the data in the sample space along with
     contextual and collective anomalies based on the time space representation since
     for fraud detection not only the value but also the frequency is an important factor.
    Whilst a certain price may be of regular value in a certain context, it may be the case that a very frequent 
    request implies a D.o.S. (Denial of Service) attack, or conversely, close prices may
     tend come together in bundles (as this is the current price of the market)
     where as, a single request followed by long 'silent' periods may be a sign of a fraud.

     \section{Related Work}
     \subsection{Anomalies in Time-series Data}
     An anomaly in time-series data is a data point (or points) at a certain time step (or steps) 
     that shows unexpected behaviors differing significantly from the behavior of previous time steps.
     Anomaly detection can be categorized in the follwing way \cite{anomaly1}:
     \begin{enumerate}
     \item Pointwise anomalies. Also known as global outliers,
      these points lay outside a user defined sensitivity parameter over the entirety of a data set.
     This user defined threshold is used to balance between type $1$ and type $2$ statistical errors.
     \item Contextual anomalies. Also referred to as conditional outliers,
      these anomalies have values that significantly deviate from other data points
       that exist in the same context (usually a period) but are not significant in the global sense. 
        The value exists within global
          expectations but may appear anomalous within certain seasonal data patterns.
     \item Collective anomalies, when a subset of \emph{continuous points} within a set is anomalous
      to the entire dataset.
       In this category, individual values are not anomalous globally or contextually
       but only the entire subset.
    Individual behavior may not deviate from the normal range in a specific section,
    but when combined with other sections these anomalies become apparent.
    \end{enumerate}
          
    \subsection{Anomaly Detection Approaches} \label{sec:det} 
    Another way to categorize anomalies is based on the implementation method: 
    
    \subsubsection{Statistical Models.} Statistical models generate statistical measures,
  such as mean, variance, median, quantile, kurtosis, skewness, and many more.
   With the generated model, a newly added time-series data can be inspected to determine
    whether it belongs to the normal boundary \cite{MARKOU20032481}.
        
%
%
   \subsubsection{Predictive Models.}  Predictive models are among the common approaches to anomaly detection.
    These methods forecast future states based on past and current states.
     We can deduce the anomaly according to the severity of the discrepancy between the    
     predicted value and the real one.
For example, autoregressive integrated moving average (ARIMA) \cite{doi:10.1080/01621459.1970.10481180}
 is frequently employed to forecast time-series.
 
  The ARIMA model is composed of three parts:
  \begin{enumerate}
\item An auto-regressive (AR) component which is composed of a weighted sum of lagged values,
 and can model the value of a random variable $X$ at time step $t$ as:
 \begin{equation}
AR(P) :  X_t = \phi_1 X_{t-1} + \phi_2 X_{ t-2} + \ldots + \phi_p X_{t-p} + \epsilon_t
 \end{equation}  
        
   where $\{\phi_i\}$ are autocorrelation coefficients and $\epsilon$ is white noise.
The parameter  $p$ is the order of AR model.
\item A moving-average (MA) component which computes the weighted sum of lagged prediction errors and is formulated as:
\begin{equation}
MA(q) : X_t = \epsilon_t - \theta_1 \epsilon_{t-1}- \theta_2 \epsilon_{t-2} - \ldots -- \theta_q \epsilon_{t-q}
\end{equation}

where $\{\theta_i\}$ are moving-average coefficients, $\epsilon_t$ denotes a model prediction error at time step $t$,
 and $q$ is the order of MA model.
\item An integrated component representing the time-series using differences, and
thus a data point at time step $t$ is $\hat X = X_t - X_{t-1}$
where $d$ denotes the order of differencing.
\end{enumerate}        
        
The differencing process makes the time-series stationary,
 resulting in ARIMA being very effective for non-stationary time-series.
If the time-series data has a seasonal variation, we can use a
variant called seasonal ARIMA (SARIMA) \cite{doi:10.1061/(ASCE)0733-947X(2003)129:6(664)}.
   In this case, we introduce the additional parameters: P, D, and Q, which deal with the seasonality.
    These parameters are used in the same manner as p, d, and q.      
    
    The parameters for the S/ARIMA can be learned in a supervised or non-supervised manner.
    In the non-supervised manner the order of $p$ and $q$ can be determined by using
     the sample autocorrelation function or via the Akaike information criterion.
      
      In the supervised manner the parameters are learned from a training set using cross-validation.
      We tested both approaches in our experiments, and saw very similar results.
      
Another commonly used algorithm is the season-trend decomposition better known as STL.
STL is a statistical method for breaking down time-series data into three different
  uncorrelated components that contain seasonality, trend and residues.
The trend analysis of the data shows a general direction of the overall data,
  while the seasonal analysis presents a pattern
  which is repeated at a fixed time intervals.
   The residue (noise) is the random oscillation or unexpected change.
   The frequency meta parameter for this algorithm can be found either by using the autocorrelation function 
   in the time 
   domain or by calculating the bandwidth of the Fourier transform in the frequency domain.

      In the category of predictive models we also include the SVM regressor,
       the nearest neighbor regressor and neuralnets based regressors.
       These methods assume a 'natural' trade line to the dataset behavior and exclude points
       which are out side a certain 'strip' of the predicted line.
       These are usually  global or contextual anomalies.
       
\subsubsection{Clustering Models.} 
Clustering based methods are choices for grouping the data into disjoint groups.
 Once a time-series is mapped into a multidimensional space, 
 clustering algorithms group them close to a cluster depending on their similarities.
  Here we assume that there is some metric in which closely related points
are close to each other and the outliers are far away.
They can vary from centroid based clustering like k-means to hierarchical based clustering like BDSCAN and others.
These algorithms are usually collective anomalies and are represented in the time (not sample) domain.
Popular data clustering methods include the k-means algorithm \cite{MacQueen1967},
 one-class support vector machine (OCSVM) \cite{DBLP:journals/jmlr/ManevitzY01},
  or Gaussian mixture model (GMM) \cite{McLa1988}.
  
OCSVM \cite{DBLP:journals/jmlr/ManevitzY01} is an unsupervised algorithm based on dividing the points 
into disjoint spaces with the margin between spaces being as wide as possible.

Instead of working in feature space it is often better to work in the kernel space instead.
Working in kernel space provide two principal benefits.
The first being that it implicitly induces a non-linear feature map,
 which allows for a richer space of classifiers. 
 The second is that, when the kernel trick is available,
it effectively replaces the dimension of the feature space
 with the size of the sample space which allows for faster computations.
Among the most popular kernels are linear kernel, polynomial kernel and
radial basis kernel (RBF).

  \subsubsection{Dimensionality Reduction Models.}
Dimensionality reduction models assume that a large scale system can be represented using a few significant factors.
 Thus, by extracting the main features via
  dimensional reduction we can reconstruct the data and separate the expected data from noise.
   Linear algebra based methods include principal component analysis (PCA)
   and singular value decomposition (SVD) \cite{DBLP:journals/corr/abs-2110-04352}.
   
   Tree based methods include isolation forest and  neural network based method include the neuralnet AutoEncoder (AE).
Isolation forest is an algorithm that detects anomalies by the depth of the tree needed for isolating each point.
At the basis of the algorithm there is an assumption that outliers are easier to
separate from the rest of the points, compared to ordinary points. 
 %
%
 In principle, anomalies are further away in the feature space and  
fewer splits are required than a normal point.
Anomalous points will therefore be defined as points where the number of partitions required until 
they are insulated is low.

  \subsubsection{NeuralNet Based Models.}
  Using deepnets can come in some of aforementioned techniques i.e. either as a predictive model or as a 
  dimensionality reduction model.
  In time-series applications, the temporal context should be considered when modeling the series.
  For this reason two general kinds neural net are considered
  \begin{enumerate}
  \item RNNS.
    RNNs have been extended with other variants, such as LSTM \cite{DBLP:journals/neco/HochreiterS97} 
    and GRU \cite{DBLP:conf/emnlp/ChoMGBBSB14}.
LSTM and GRU address the vanishing or exploding gradient problem,
 where the gradient becomes too small or too large as the network goes deeper. 
 There are multiple gates in an LSTM and a GRU cell,
  and they can learn long-term dependencies by determining the number of previous states
   to keep or forget at every time step.
 Meanwhile, the dilated RNN is proposed to extract multiscale features 
    while modeling long-term dependencies by using a skip connection between hidden states.
     Shen et al. \cite{NEURIPS2020_97e401a0} for instance, adopt a three layer dilated RNN and extract features from each layer 
     to jointly consider longterm and short term dependencies.
RNN-based approaches are generally used for anomaly detection in two ways.
 One is to predict future values and compare them to predefined thresholds
  or the observed values.
   This strategy is applied in \cite{NEURIPS2020_97e401a0,DBLP:conf/kdd/HundmanCLCS18,DBLP:journals/corr/abs-1711-00614,DBLP:conf/ijcai/KieuYGJ19}.
   The other is to construct an autoencoder in order to restore the
    observed values and evaluate the discrepancy between the reconstructed value and observed one.
     This strategy is used in \cite{DBLP:conf/soca/HsiehCH19,DBLP:conf/icann/LiCJSGN19,DBLP:conf/icann/LiCJSGN19,DBLP:conf/kdd/SuZNLSP19,DBLP:conf/acml/GuoLWYJL18}.
     
  \item CNNs.  
   Although the RNN is the primary option for modeling time-series data,
    CNN sometimes shows better performance in several applications that work with short term data. 
     \cite{DBLP:conf/bigcomp/ChoiLCK20,DBLP:journals/corr/abs-1905-13628,DBLP:conf/ijcai/ZhouLHCY19}
   
    By stacking convolutional layers, each layer learns a higher level of features.
     In addition, the pooling layers introduce non-linearity to the CNN,
      allowing them to capture the complex features in the sequences.
Instead of explicitly capturing the temporal context,
 the CNN models learn patterns in segmented time-series.
  
   \end{enumerate}

  In the field of the cryptocurrency anomaly detection many previous works were done on finding
  anomalies using machine learning techniques.
 Bartoletti et al. \cite{DBLP:journals/fgcs/BartolettiCCS20}, Chen et al. \cite{DBLP:conf/www/ChenZCNZZ18},
  \cite{DBLP:journals/access/ChenZNZZ19} and Jung et al. \cite{DBLP:conf/blockchain2/JungTGG19} analyzed Ponzi schemes 
of Ethereum trade developed as smart contracts.
 
Chen et al. \cite{DBLP:conf/www/ChenZCNZZ18} used a regression tree model (XGBoost)
 to detect Ponzi schemes on Ethereum.
A subsequent work of the same authors \cite{DBLP:journals/access/ChenZNZZ19} 
improved the results by using the Random Forest (RF) classifier.
 Jung et al. \cite{DBLP:conf/blockchain2/JungTGG19}  analyzed both behavior and code
  frequency of contracts to detect Ponzi schemes,
  and obtained a precision of $99\%$ and recall of $97\%$.
    
   Bartoletti et al. and Toyoda et al. used addresses clustering techniques
    \cite{DBLP:conf/socialcom/ReidH11},\cite{DBLP:journals/cacm/MeiklejohnPJLMV16} 
 to extend their datasets of scam addresses,
    and both applied \emph{supervised} machine learning techniques to detect Ponzi schemes.
     In particular, Toyoda et al. used XGboost and Random Forest classifiers,
    and Bartoletti et al. used a cost-sensitive Random Forest classifier,.

         Schnaubelt   \cite{DBLP:journals/eor/Schnaubelt22}, Shahbazi \cite{DBLP:journals/access/ShahbaziB21a}
         and Tanwar\cite{DBLP:journals/access/TanwarPPPSD21} showed the use 
   of deep reinforcement learning to find the optimal placement of cryptocurrency limit orders and
     Akba et al. \cite{DBLP:journals/access/AkbaMGA21} used predictive models for detection
      of a manipulator in the cryptocurrency market.

\section{Experiments}
For our experiments we ran all the above mentioned algorithms and compared the results for anomalies
in the same category. 
In the category of predictive models we used ARIMA, SARIMA, STL, LSTM based regressors,
kernel ridge regression, support vector regressor (SVR) RBF, SVR with a polynomial kernel and nearest neighbor regressor.
In the category of dimensionality reduction models we used Isolation forest, LSTM AE and PCA based models.
In the category of cluster based models we used the OCSVM, k-means and DBSCAN based model.

The datasets for these experiments are in the public domain, collected from etherscan.io and accessible through public APIs.
The name of stock markets and the active accounts are available on our GitHub.
Also comprehensive results of the different algorithms with their accuracy can be found on GitHub \footnote{ \url{https://github.com/erankfmn/anomaly\_detection}}.
Here we present some illustrative results, while the conclusion section summarizes the result from all datasets and algorithms.
The algorithms are briefly explained in subsection \ref{sec:det}.

\begin{figure}[!htbp]
\centering
\caption{example of point summation for time representations}
\includegraphics[width=\textwidth,height=\textheight,keepaspectratio]{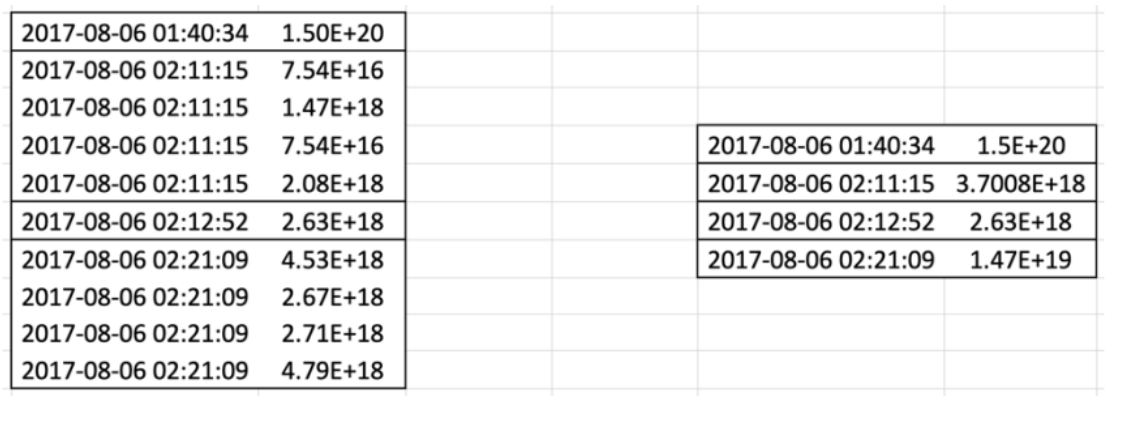}
\label{example_of_summation}
\end{figure}

%
%

\subsubsection {Processing the Data.}
The data is given as the dates and amounts of the ethereum transaction for a single account.
Dates and time where no transaction took place are not given in the dataset (since this is the common case).
While some anomaly algorithms work better in the sample space and even may be
confused by a long period of null transactions (essentially, marking every transaction as an anomaly).
Other algorithms which search for an anomaly of a point in temporal space need their representation in time domain instead.

    When treating the series based on time rather than on samples an issue arrises  
    as the data is typically given in the format of value and date, it maybe the case 
    that some exact dates cooccur (at the exact same second).
     While working in the sample domain these points
    are regarded as different points however, in the time domain it is unclear as to how they should be
     treated since there can only be one data point for each sampling time.
     We decided in these cases to merge these points into to a single point.
      Other possible approaches may include taking the maximum or the average value,
    but after exhaustive trial and error, we found that merging was more appropriate
     in the context at hand.
     An example of such a manipulation is illustrated in figure \ref{example_of_summation}.

\begin{figure}[!htbp]
\centering
\caption{Raw Data of the payment amount gas price and gas limit}
\includegraphics[width=\textwidth,height=\textheight,keepaspectratio]{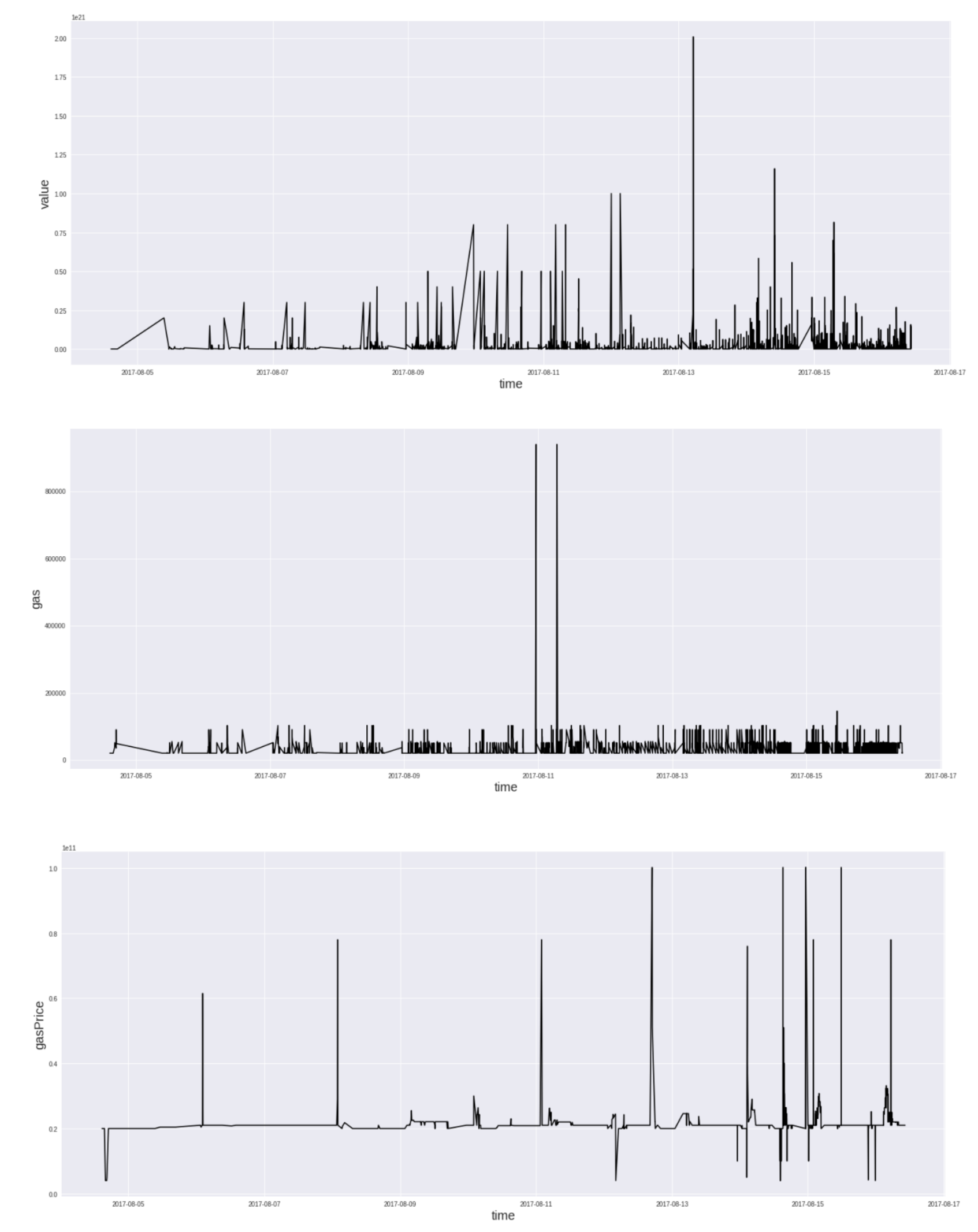}
\label{Raw Data}
\end{figure}

Figure \ref{Raw Data} shows the raw data in the sample space for the fields of payment amount, gas limit and gas price.
It is apparent that the different fields are 
highly uncorrelated and demonstrate different anomaly behavior.

\begin{figure}[!htbp]
\centering
\caption{ARIMA SARIMA STL results for anomalies in the payment field}
\includegraphics[width=\textwidth,height=\textheight,keepaspectratio]{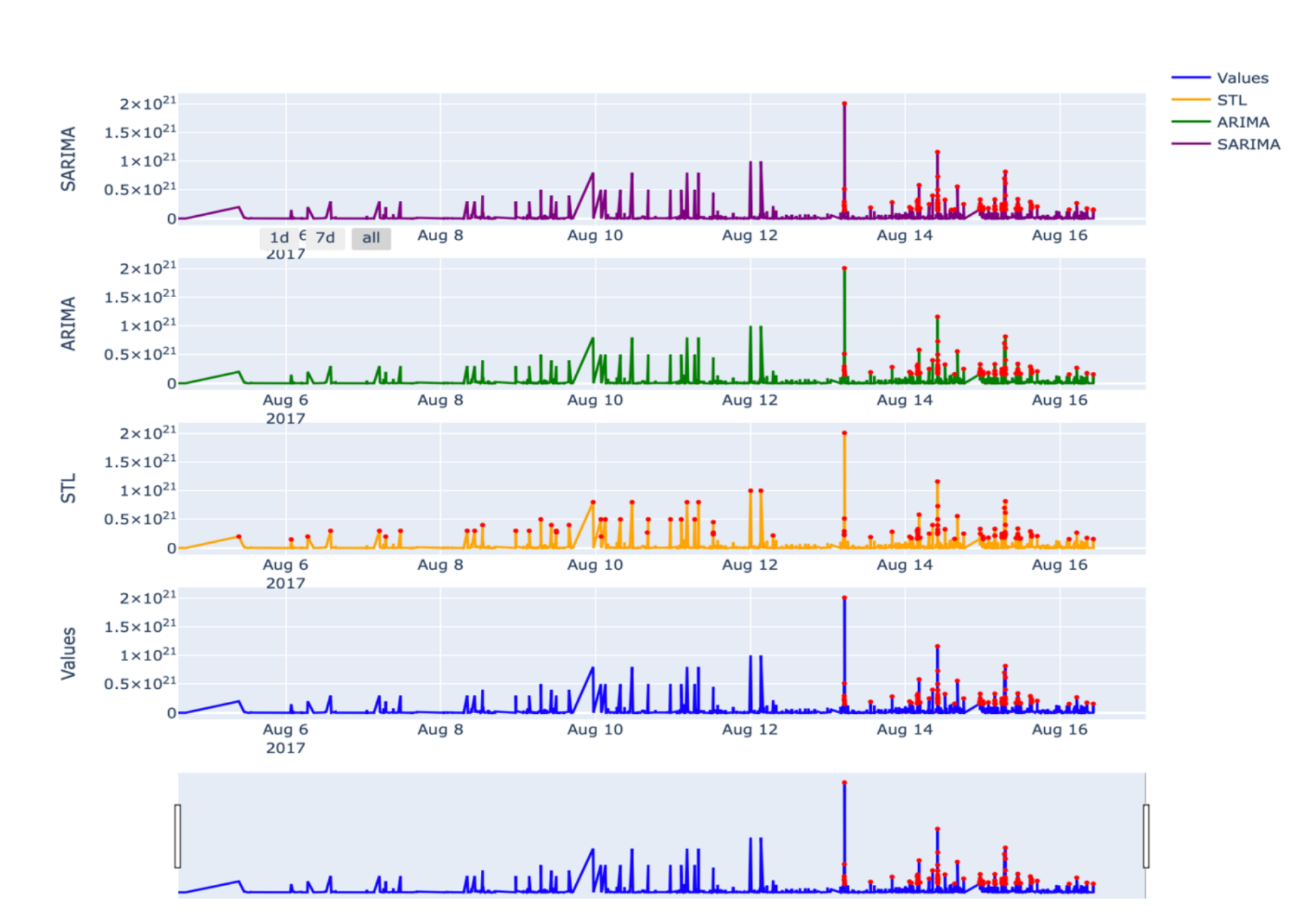}
\label{STL res}
\end{figure}

Figure \ref{STL res} shows the concurrent result of  ARIMA, SARIMA and STL algorithms
for the payment amount feature and at the bottom of the graph shows points which more than two algorithms concur.

In order to estimate the parameters for S/ARIMA  cross validation was used, the data was split into train 
and test $70\%-30\%$ and validated over all possible combinations of parameters over the range of $1-3$. 
Test values whose difference between the prediction and actual value greater than
 $3 \cdot RMS$ were detected as an outlier. 
  
 We can concluded that the three algorithms identified the same samples
  quite similarly. For example, for SARIMA accuracy is $97\%$ and recall $96.6\%$ .
The ARIMA and SARIMA algorithms identified almost exactly the same anomalies, probably due
  to the fact the that the seasonal component did not have significant weight in the sampling classification.
  
 The STL algorithm produced similar result to the two other algorithms.

 The points that were identified were usually points at the beginning or end of major trend changes.
 
\begin{figure}[!htbp]
\caption{ARIMA SARIMA STL results for anomalies in the gas fields}
\subfloat[Gas Price]{\includegraphics[width = 3in]{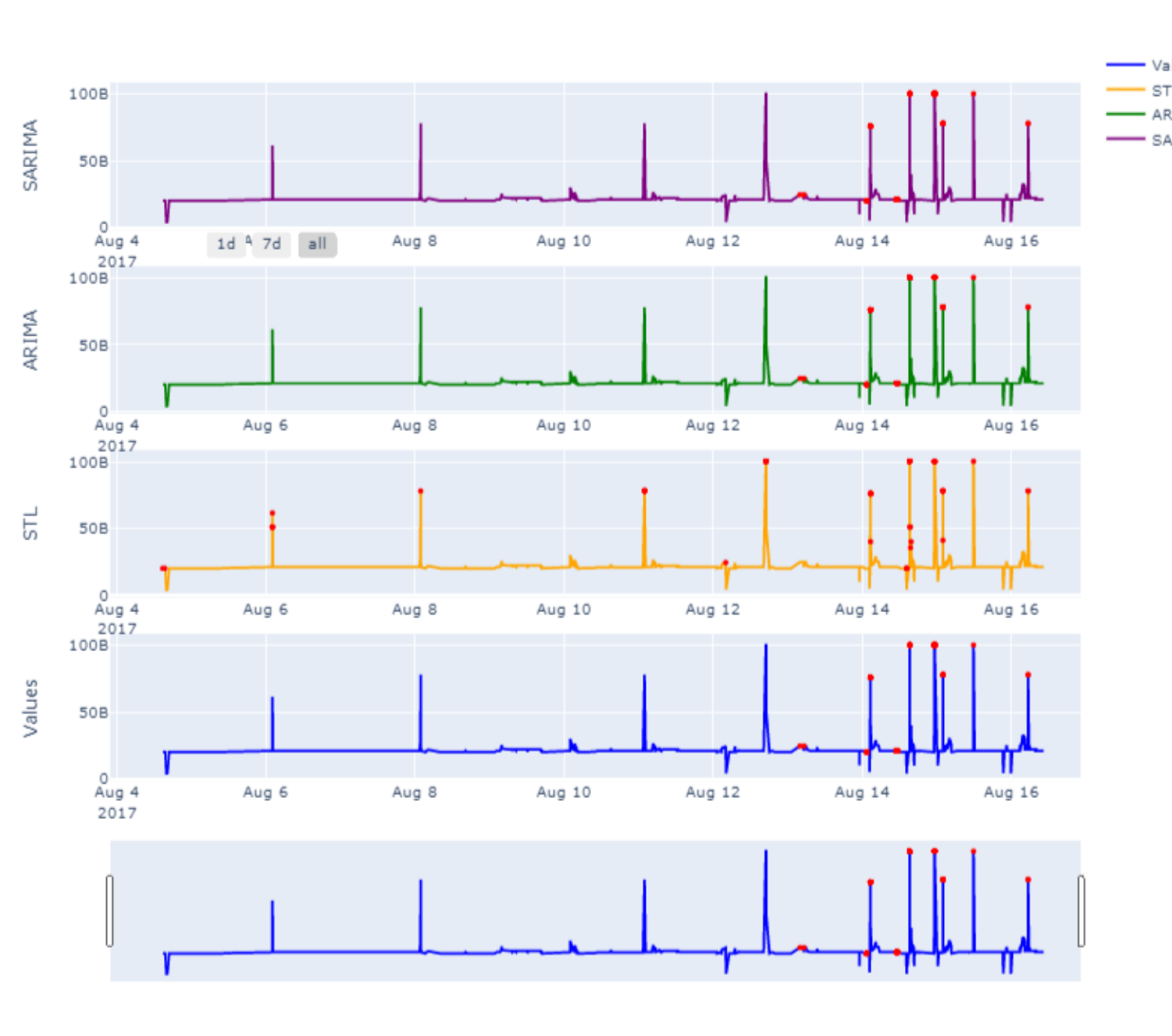}} 
\subfloat[Gas Limit]{\includegraphics[width = 3in]{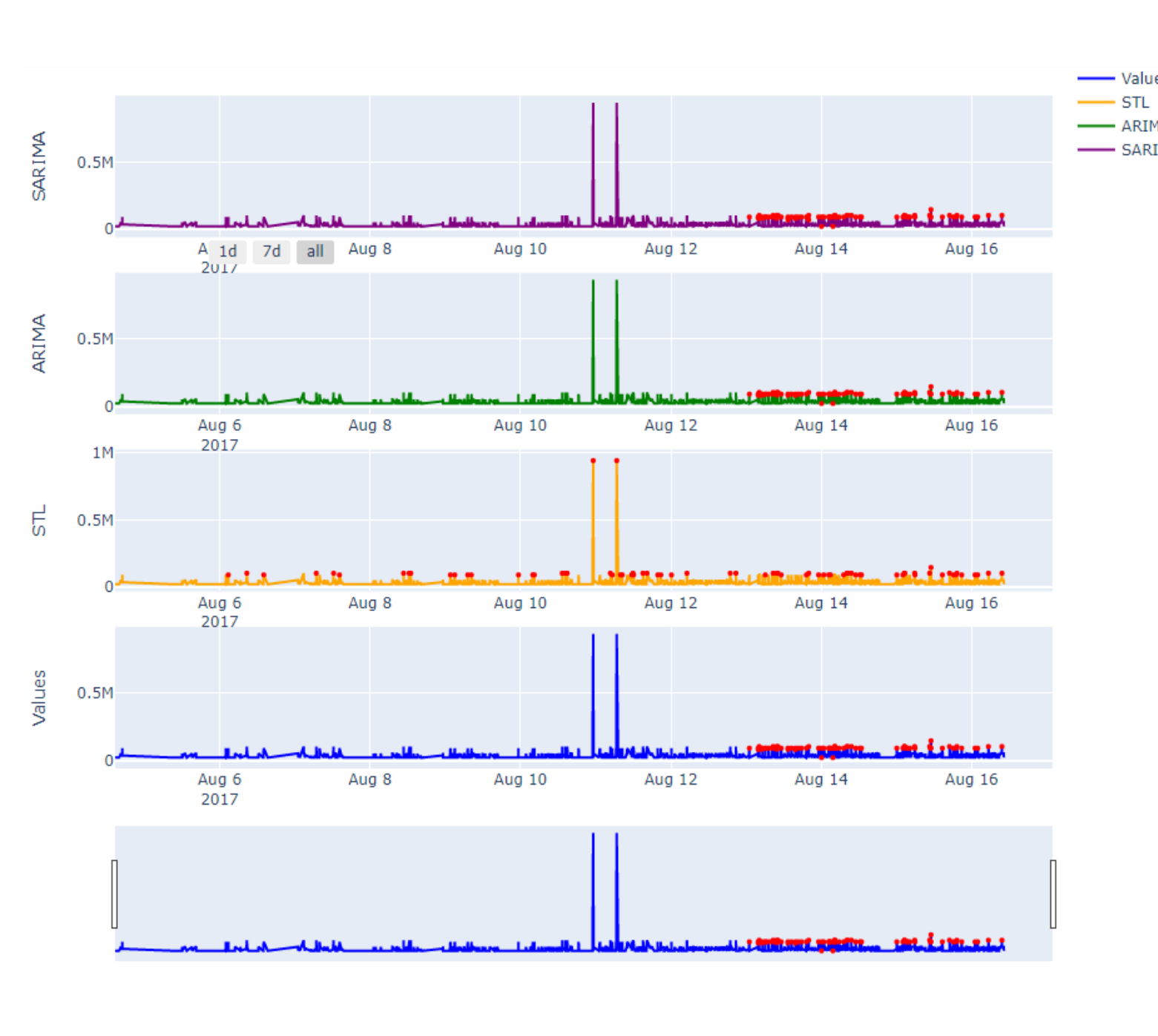}}
\label{gas stl}
\end{figure}

The two following graphs in figure \ref{gas stl} are for the gas price and gas limit yielding the same results.

\begin{figure}[!htb]
\centering
\caption{Anomalies detected for payment by regression trees and LSTM}
\includegraphics[width=\textwidth,height=\textheight,keepaspectratio]{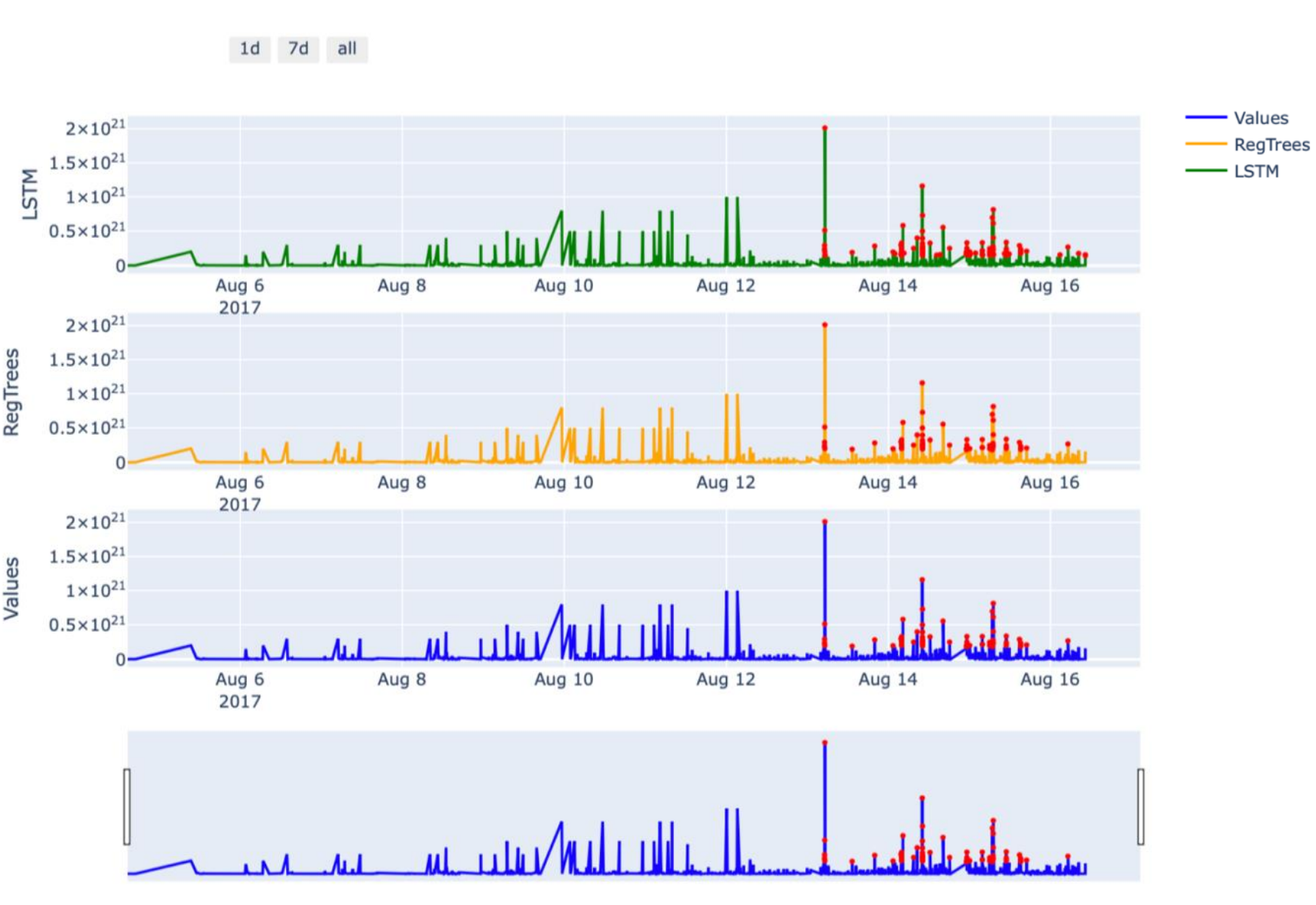}
\label{payment reg}
\end{figure}

\begin{figure}[!htbp]
\caption{Anomalies detected for the gas fields by regression trees and LSTM}
\subfloat[Gas Price]{\includegraphics[width = 3in,height=2in]{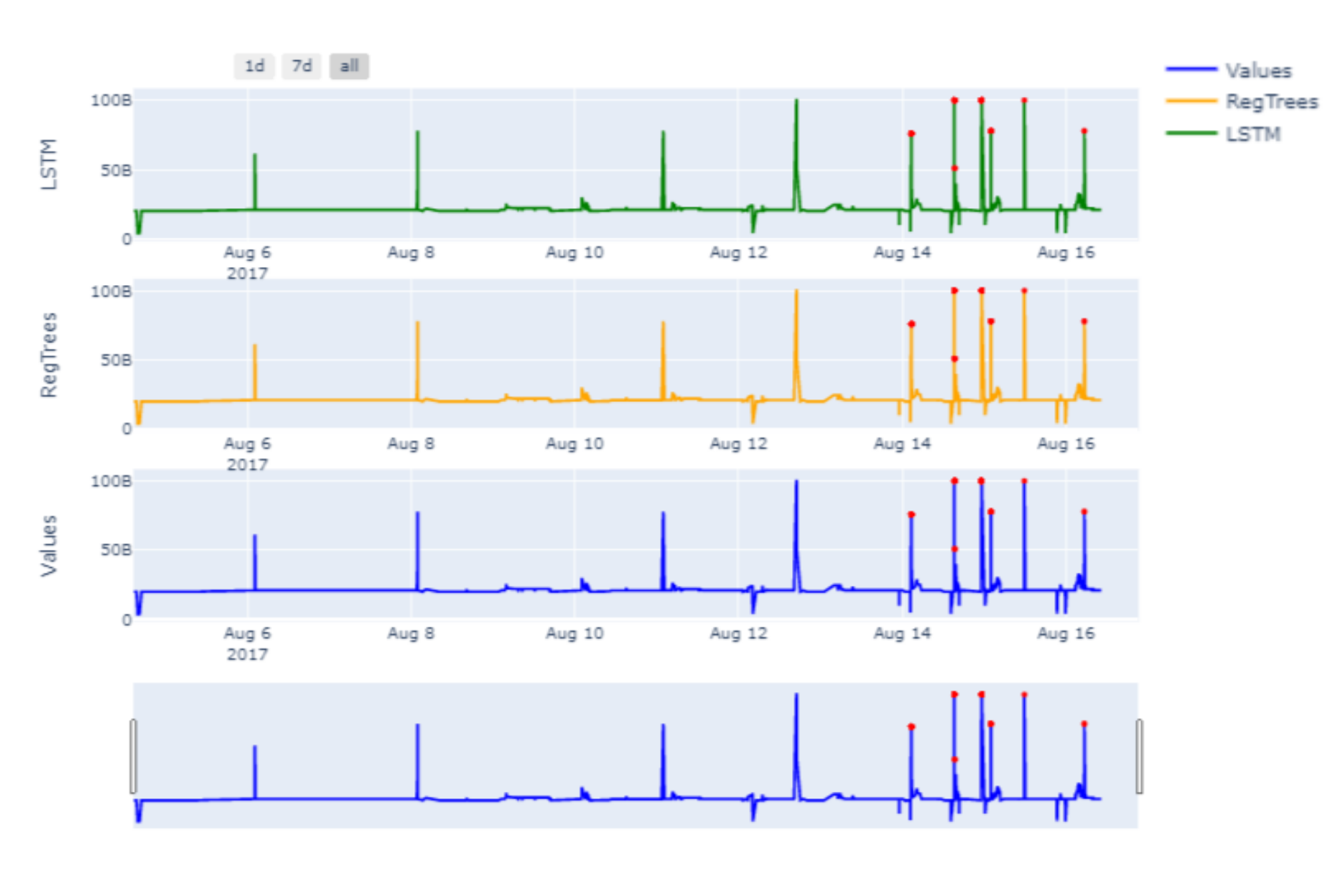}} 
\subfloat[Gas Limit]{\includegraphics[width = 3in]{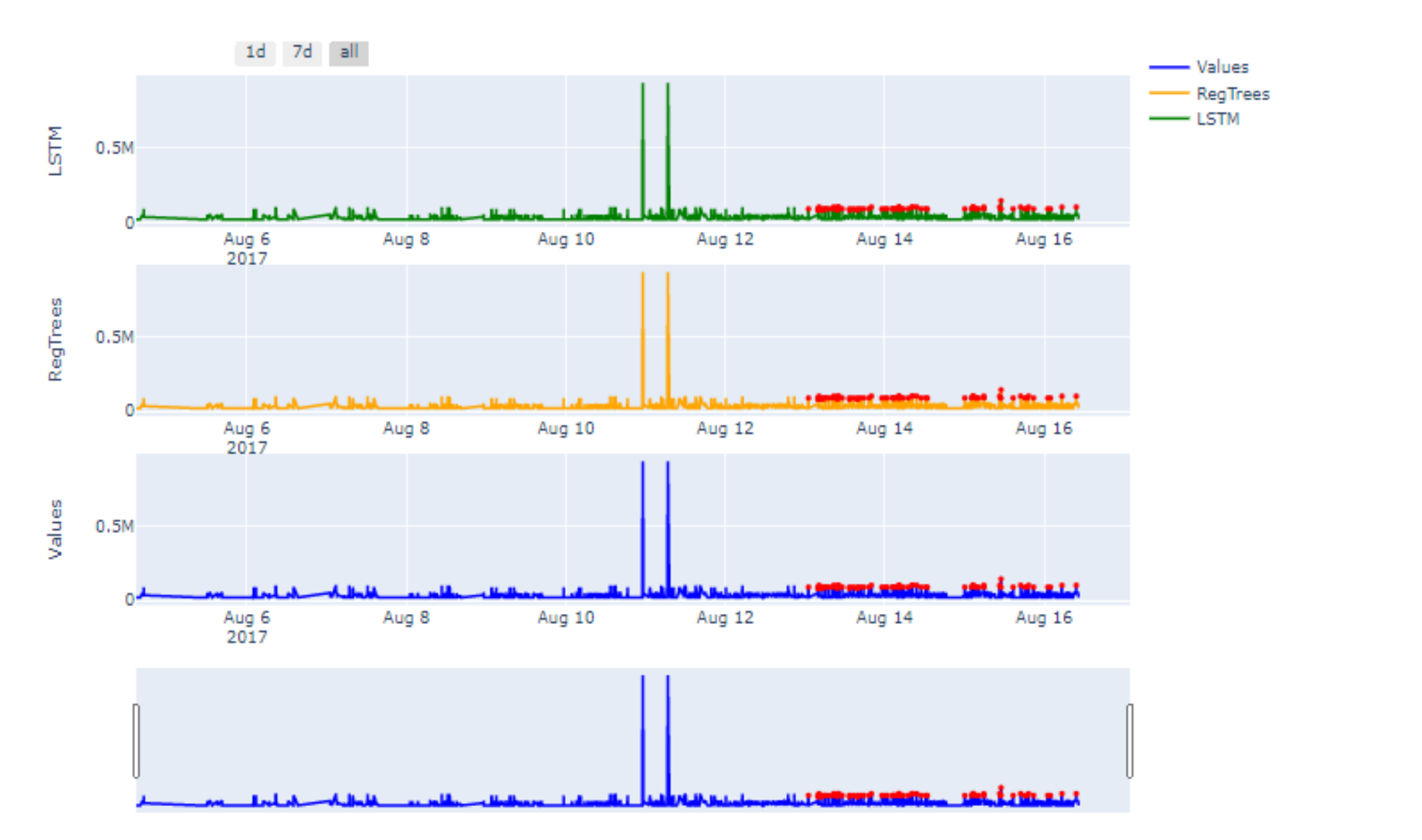}}
\label{gas reg}
\end{figure}

Figure \ref{payment reg} shows the results
of LSTM and regression trees for the payment value.
Figure \ref{gas reg} shows the same result for gas price and gas limit. 
Since predictive models are supervised models they are split to train and test sets.
When commerce start to rise the tendency of predictive models is to keep the tendency line rising
and when sudden and abrupt changes occur they are classified as anomalies. 
The LSTM architecture was constructed by 3 layers of LSTM consisting of $128,64,32$ units respectively,
followed by a dropout layer - this is the encoder part.
Afterwards a repeated vector layer followed by the decoder unit of 3 layers of LSTM $32,64,128$ units respectively.

\begin{figure}[!htbp]
\begin{center}
\centering
\caption{Anomalies detected for payment by isolation forest and OCSVM}
\includegraphics[width=\textwidth,height=\textheight,keepaspectratio]{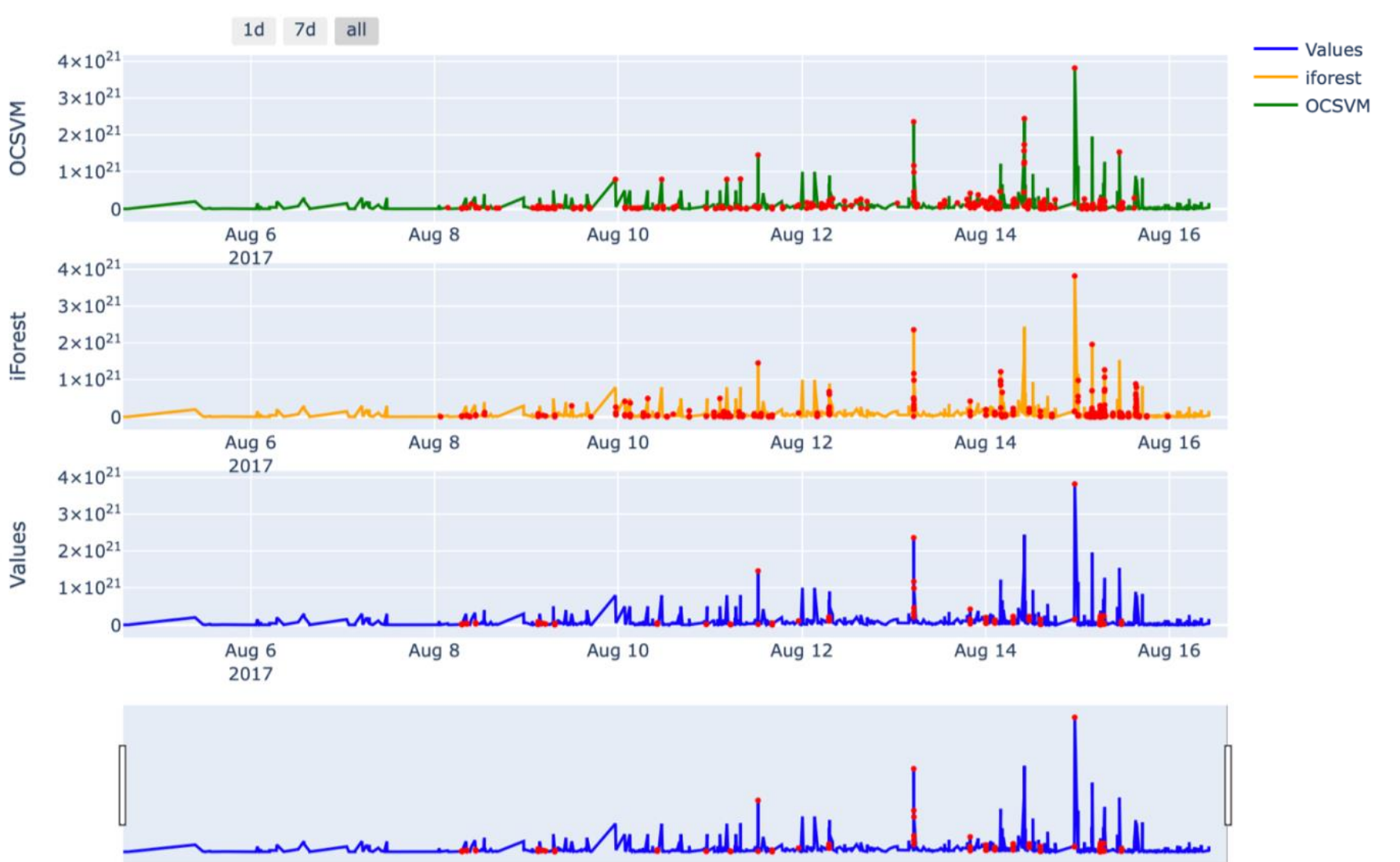}
\label{payment iso}
\end{center}
\end{figure}

\begin{figure}[!htbp]
\caption{Anomalies detected for gas fields by isolation forest and OCSVM}
\subfloat[Gas Price]{\includegraphics[width = 3in,height=2in]{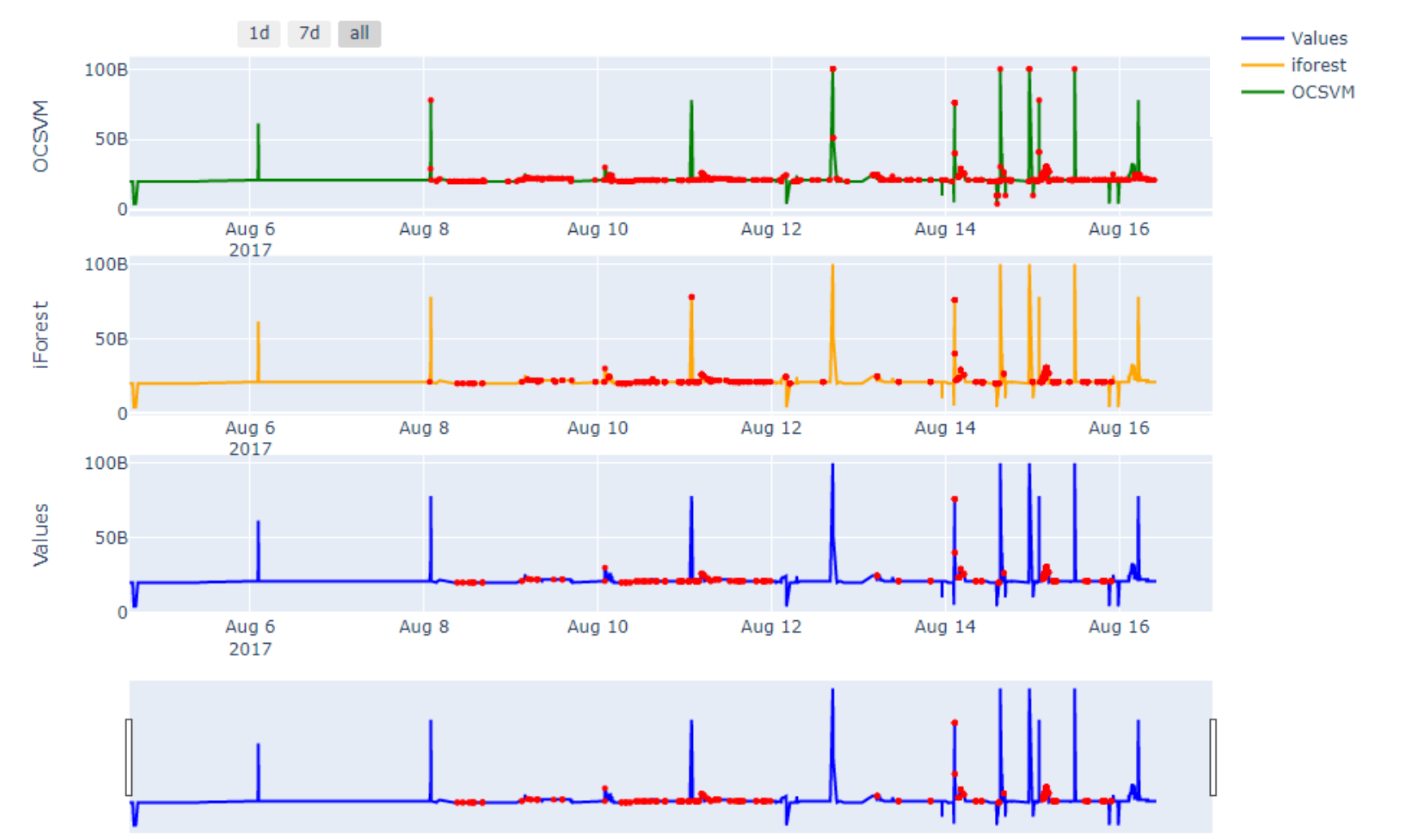}} 
\subfloat[Gas Limit]{\includegraphics[width = 3in,height=2in]{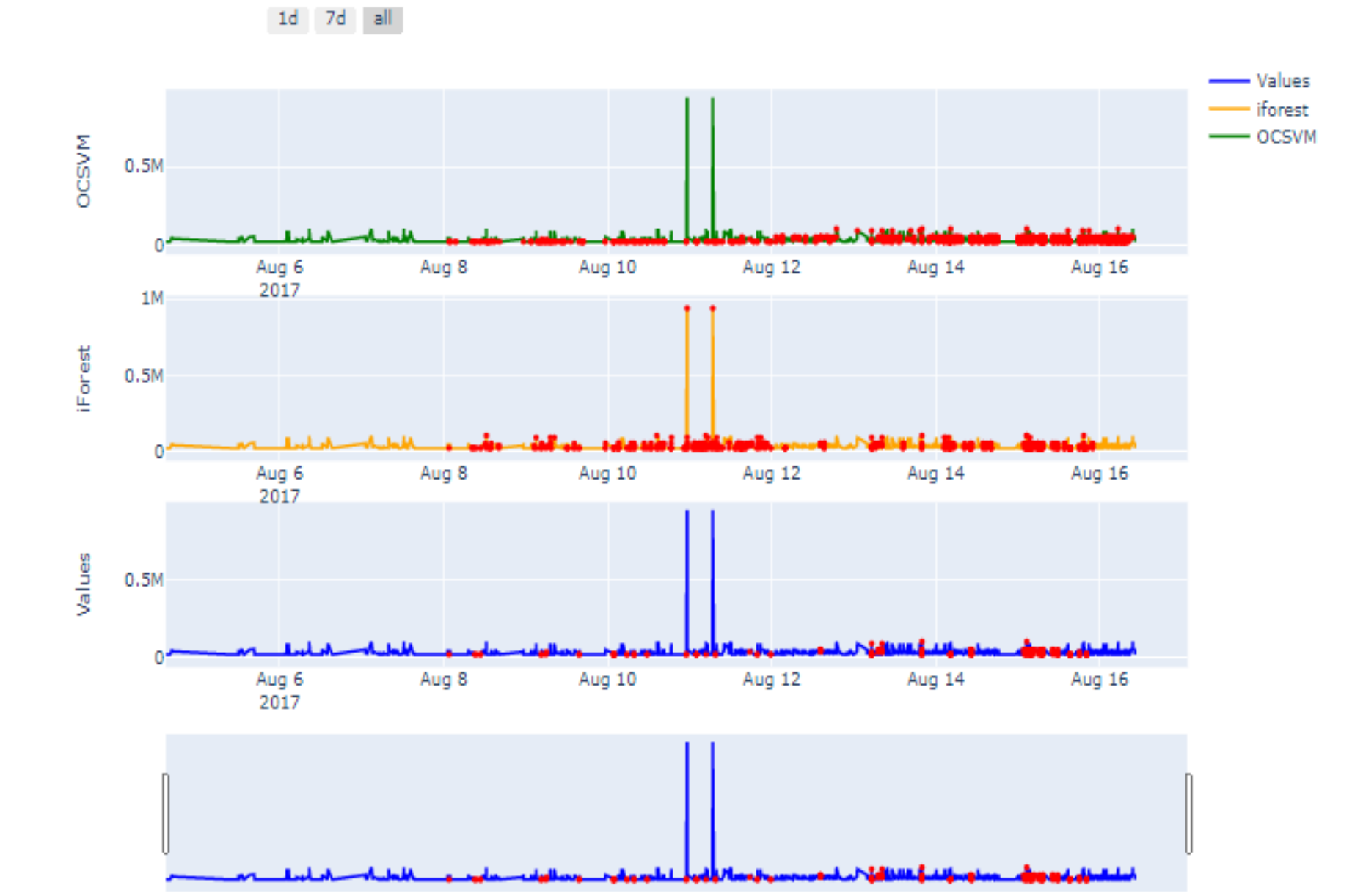}}
\label{gas iso}
\end{figure}

Figures \ref{payment iso} and \ref{gas iso} were taken from algorithms who look at collective anomalies.
For the isolation forest and the OCSVM collective points 
were prepared as a continuous time-series, the
sliding window was set to be $5$ continuous minutes 
and each sampling point can belong to several vectors at the same time in the time domain.
These type of algorithms perform online learning, they receive
 a set of samples of about five minutes and compares them to the previous four day database.
The database is also updated every four days as detection needs to happen in real time. 
 
 We used OCSVM with RBF kernel.
 The regularization parameter $C$ was $5$-fold cross-validated over the set $\{2^{5}, 2^{3}, \ldots, 2^{15}\}$,
  and for the RBF kernel, the $\gamma$ parameter was five-fold cross-validated over the set
   $\{2^{5}, 2^{3}, \ldots, 2^{3}\}$.\footnote{This is also true for the SVR where also tried 
   both polynomial and RBF kernels}

\section{Conclusion and Further Research}
In this paper we evaluated the anomalistic behavior of Ethereum trade of several accounts 
for a real time fraud detector. The detection searched for pointwise as well as contextual and
 collective anomalies both in the sample domain, the time domain and the frequency domain.
We used many well known algorithms traditional and new
 to try and compare their performances.
We categorized the algorithms according to the their strategy of detection.
 We marked a point as an anomaly
  when there was 
 a majority vote on that point.
 In the category of predictive models we found that classical as well as neural based methods have
 an accuracy of $97\%$ and a recall of $96.6\%$.
  They work well together and tend to appear when trade line are changes abruptly. 
 Dimensionality reduction methods on the other hand are contextual and collective anomalies and work
 well both in time or sample
 representations.
Here we found poor results since especially neuralnets AE detected more points than other methods such as 
 isolation forest and PCA and we believe them to be more sensitive due to their complexity.
 Clustering methods have an accuracy of $96.3\%$ and a recall of $93.2\%$.
 They are collective anomalies that work well both in the time domain as well as in the sample domain.
 They also work both as univariate and multivariate models detecting uncorrelated vectors.
 All model give reliable real time results either pretrained or online learners.
 The over all conclusion of our research is that for a real time anomaly detector
 with unsupervised data the best approach is
 to have parallel processing divided according to the anomalistic behavior:
 pointwise local or global, contextual, collective,in the sample or time representation.
 And have them work simustansouly while using a majority vote.
Models from the same category traditional and new tend to have a high rate of agreement
 and are therefore reliable and relatively fast for real time fraud detection.

\pagebreak

%
%
%
\bibliographystyle{splncs04}
\bibliography{anamoly}
%
%
%
%
%
\end{document}